\documentclass[]{interact}

\usepackage{epstopdf}
\usepackage[caption=false]{subfig}

\theoremstyle{plain}

\theoremstyle{definition}

\theoremstyle{remark}

\usepackage{subfig}
\usepackage{setspace}
\usepackage{float}
\usepackage{graphicx}
\usepackage{amssymb}
\usepackage{tikz}
\usetikzlibrary{shapes.geometric,arrows}
\tikzstyle{startstop} = [rectangle, rounded corners, minimum width=3cm, minimum height=1cm,text centered, draw=black, fill=red!30]
\tikzstyle{io} = [trapezium, trapezium left angle=70, trapezium right angle=110, minimum width=3cm, minimum height=1cm, text centered, draw=black, fill=blue!30]
\tikzstyle{process} = [rectangle, minimum width=3cm, minimum height=1cm, text centered, draw=black, fill=orange!30]
\tikzstyle{decision} = [diamond, minimum width=3cm, minimum height=1cm, text centered, draw=black, fill=green!30]
\tikzstyle{arrow} = [thick,->,>=stealth]

\begin{document}

\articletype{ARTICLE TEMPLATE}

\title{An XGBoost-Based Forecasting Framework for Product Cannibalization}

\author{
\name{Gautham Bekal and Mohammad Bari}
\affil{T-Mobile, Two Newport, 3625 132nd Ave SE, Bellevue, Washington, 98006, USA}
}

\maketitle

\begin{abstract}
Two major challenges in demand forecasting are product cannibalization and long term forecasting. Product cannibalization is a phenomenon in which high demand of  some products leads to reduction in sales of other products.
Long term forecasting involves forecasting the sales over longer time frame that is critical for strategic business purposes. Also, conventional methods, for instance, recurrent neural networks may be ineffective where train data size is small as in the case in this study. This work presents XGBoost-based three-stage framework that addresses product cannibalization and associated long term error propagation problems. 
The performance of the proposed three-stage XGBoost-based framework is compared to and is found superior than that of regular XGBoost algorithm.
\end{abstract}

\begin{keywords}
Custom Objective Function, XGBoost, Product Cannibalization, New Product Introduction(NPI)
\end{keywords}

\section{Introduction}

Demand forecasting is an important use case in supply chain, inventory management, retail, etc. Predicting the sales of various products within the portfolio is crucial for making business decisions and other downstream tasks such as inventory optimization and target calculation, etc. One of the major challenges in making such forecasts is taking the effect of product cannibalization into account. Product cannibalization occurs when demand for a certain product within the portfolio increases that may be due to launch of a new product. This consequently reduces the sales of older products. This interaction between different data samples leads to the fact that total demand of all products remains stable but with large variations in the demand of individual products within the portfolio.

Machine learning allows us to model complex dynamics and capture large number of input variables over traditional statistical models. Generally, machine learning models try to optimize the cost function by using input features to the model and updating the model parameters accordingly. However, in product cannibalization the demand of a given product is being impacted by the demand of a different product that is not a part of the input feature set.
In this work, the proposed framework is to make accurate sales forecast of old products that are cannibalized due to launch of newer products.
Consequently, off the shelf machine learning models like XGBoost or neural networks are not able to capture the interactions between different data samples during the training.

The work presented in ~\cite{Zara} focuses on the product demand forecast for the distribution team, and therefore only tackles one
week ahead demand forecast. One of the important work ~\cite{Remy} assumes short sale cycle for a given product and hence is useful in making short term forecasts. We tackle both short term and long term cannibalisation.
 Also, we have focused on long term forecasting for 8 or more weeks. The work ~\cite{Zara} uses multinomial model hence output is discrete, i.e. it is for categorical output. Our use-case focuses on continuous variable since we are trying to forecast product sales that is a real number.

The second major hurdle in the above use case is when making long term predictions under limited training data. One of the standard techniques is to use Recursive Multi-step Forecast ~\cite{ML_mastery} technique where prediction at a given time-step is used as an input feature to make forecast for next time step. The problem with this approach is that it causes error propagation when making long term forecast. 

On the other hand, ~\cite{Hossein} provides with multiple approaches to modelling long term forecast. However, these are standard approaches that do not cater our use case completely. The reason being, the paper's standard approaches do not take cannibalization into account and thus the performance is severely deteriorated. 
 
The paper ~\cite{Carlos} deals with cannibalization due to promotional impact whereas in our use case we are concerned with cannibalization due to new product launches. In both cases however the underlying assumption is on casual relationship between cannibalized and cannibalizing product.

Our work improves the forecast accuracy of old products whose sales have been affected by the addition of newer products to the portfolio. In this study any product after 4 weeks of launch is assumed to be an old product. Thus, all products within 4 weeks of launch period are NPI products.
The machine learning algorithm we introduce in this work handles these limitations.

\section{Data Preparation}
The dataset is a tabular data with \emph{n} rows and \emph{d}+1 columns. The rows are divided between train set, indexed 1 to \emph{m} rows, and test set, indexed \emph{m} + 1 to n rows.
The input matrix to the model is defined as \emph{X} and is in \emph{d}-dimensional space. The target variable is \emph{Y}, which is a continuous real value corresponding to number of units sold each week. We train and predict using our three-stage framework for each product category separately independent of one another. Thus for our experiment we have 3 datasets corresponding to 3 separate product categories.
Each product category contains \emph{count} as the number of products sold for a given specific week. Also, the number of products within a product category can vary week-over-week.
The general overview of the dataset is as shown in Table 
\ref{tab:Table_one}.

\begin{table}[]
\caption {Experiment Results} \label{tab:Table_one} 
\centering
\begin{center}
\scalebox{0.9}{
\begin{tabular}{|l|l|l|l|l|l|l|l|l|}
\hline
 \begin{tabular}[c]{@{}l@{}} Category\end{tabular}
 &\begin{tabular}[c]{@{}l@{}} Date\end{tabular}
  &\begin{tabular}[c]{@{}l@{}} Product \end{tabular}
   &\begin{tabular}[c]{@{}l@{}} Promotion\end{tabular}
   &\begin{tabular}[c]{@{}l@{}} Seasonality\end{tabular}
   &\begin{tabular}[c]{@{}l@{}} sale(t-3)\end{tabular}
    &\begin{tabular}[c]{@{}l@{}} sale(t-2)\end{tabular}
     &\begin{tabular}[c]{@{}l@{}} sale(t-1)\end{tabular}
    &\begin{tabular}[c]{@{}l@{}} sale(t)\end{tabular}
  \\ \hline Category\_A & Date\_1 & A\_1  & Promo\_1 & Season\_1 & 1000 & 1110 & 1150 & 1200
    \\ \hline Category\_A & Date\_2 & A\_1  & Promo\_1 & No\_Seasonality & 1110 & 1150 & 1200 & 1500
      \\ \hline Category\_A & Date\_3 & A\_1  & Promo\_2 & No\_Seasonality & 1150 & 1200 & 1500 & 1300
      \\ \hline Category\_A & Date\_4 & A\_1  & Promo\_4 & Season\_2 & 1200 & 1500 & 1300 & 2000
      \\ \hline Category\_A & Date\_5 & A\_1  & No\_Promo & Season\_3 & 1500 & 1300 & 2000 & 1650
      \\ \hline Category\_A & Date\_1 & A\_2  & Promo\_4 & Season\_1 & 2000 & 4200 & 4000 & 5000
    \\ \hline Category\_A & Date\_2 & A\_2  & Promo\_1 & No\_Seasonality & 4200 & 4000 & 5000 & 3600
      \\ \hline Category\_A & Date\_3 & A\_2  & Promo\_6 & No\_Seasonality & 4000 & 5000 & 3600 & 3900
      \\ \hline Category\_A & Date\_4 & A\_2  & No\_Promo & Season\_2 & 5000 & 3600 & 3900 & 4200
      \\ \hline Category\_A & Date\_5 & A\_2  & No\_Promo & Season\_3 & 3600 & 3900 & 4200 & 5500
                \\ \hline
\end{tabular}}
\end{center}
\end{table}

Each row corresponds to number of units the product sold in the corresponding week as given  by the column \emph{sale(t)}. Hence, \emph{sale(t)} is the target variable \emph{Y}. Every other column in above table corresponds to input feature \emph{X} which impacts the sales of column \emph{sale(t)}.
Our objective is to make accurate prediction of \emph{sale(t)} for a given product on any given date.
Since, we are implementing a forecasting model the sales of a product in a given week, i.e. \emph{sale(t)}, is affected by sales of the same product in previous weeks. Hence, we have included sales of previous three weeks of the same product as a part of the input feature set \emph{X}. For historical data when the sales is available, it is possible to fill \emph{sale(t-1)}, \emph{sale(t-2)} and \emph{sale(t-3)}. However, when making long term predictions on future dates the previous weeks sales is not available to us. For example, if we have historical sales available upto August 31, 2020 in which case when making a prediction for September  28, sales of September 07, September 14 and September 21 is not available to us. Hence, \emph{sale(t-3)}, \emph{sale(t-2)} and \emph{sale(t-1)} will be unfilled in the input matrix \emph{X}. To overcome the issue we will borrow an idea from natural language prediction. While making prediction on future dates for testing, we will use the prediction made on previous dates and append it to the columns in \emph{X}. In the above example, this would mean we would first make a prediction on September 07, 14, and 21 and append these predictions to columns of \emph{sale(t-3)}, \emph{sale(t-2)} and \emph{sale(t-1)} respectively. 
This above concept of using previous week sale prediction as an input to make a prediction on next week is referred to as back-padding.
Later, we will discuss how this widely used concept of back-padding in time-series forecasting leads to the problem of error propagation for longer time horizon during prediction.

\section{Motivation}
The standard XGBoost model for regression modelling uses SE loss function given by,
\begin{equation}\label{equation_one}
l=\frac{\sum_{i=1}^{m}(f(X_i) - Y_i)^2}{m}  
\end{equation}
where, \emph{m} is the number of training samples.
\emph{X} is an \emph{m} x \emph{d} input vector to the XGBoost
\emph{Y} is the target variable which takes on continuous real value and has dimension \emph{m} x 1
\\*\-\hspace{1cm}The XGBoost model is trained on \emph{X} to approximate the target variable \emph{Y} by a function  \emph{f}.
The usual assumption here is that the behaviour of \emph{Y} can be completely modelled using \emph{X}. However, in product cannibalization we observe \emph{$Y_i$} also depends on the values of \emph{$Y_j$} where, $i \neq j$. Here, \emph{i} and \emph{j} is the row number.
Here, for a given target data point  \emph{$Y_i$}, \emph{$Y_j$} cannot be made as a part of independent feature  \emph{$X_i$} for the following reason.
\\*\-\hspace{1cm}For testing, all the target variables \emph{$Y_{(m+1)}$} to \emph{$Y_n$} are unknown. Hence,for test dataset between \emph{m}+1 to \emph{n}  the independent variable \emph{$X$} will have certain columns empty. Hence, in the test dataset for a given target variable data point \emph{$Y_i$} on which we want to make prediction we cannot have  \emph{$Y_j$} as a part of input matrix, since \emph{$Y_i$} and \emph{$Y_j$} are both unknown. Consequently, the XGBoost model based on SE will lose important information pertaining to cannibalization and thus we need an alternative approach to incorporate the cannibalization information to the model. 
In time series forecasting to forecast over long term horizon, we depend on the predictions made by previous time periods. Hence, the models are auto-regressive.
Unfortunately, this leads to the problem of error propagation that become more severe as the prediction time horizon increases. This is true in our study also, since we are forecasting sales for 8 to 14 weeks into the future respectively. 
\\*\-\hspace{1cm}To overcome the two challenges we introduce a sum constraint based modelling technique that involves predicting for target variable data-points \emph{$Y_{(m+1)}$} to \emph{$Y_n$} and also simultaneously solving constraint based equation that incorporates cannibalization information.
The available dataset for this work is quite small with individual category containing approximately 3000 records and large number of independent variables. 
\emph{X}. Hence, feed forward or other deep learning frameworks have not been used to construct the framework.
XGBoost is an obvious algorithm of choice for implementing three-stage framework mainly because it has superior performance on small datasets and implementing custom objective functions is relatively straight forward.

\section{Algorithm}
We make a fundamental assumption that, sum of sales over the week for a given category is easily obtainable from domain knowledge and is already known to us. What we are left with is to make week over week predictions for individual products
belonging to the specific category.
Mathematically it can defined as,
\begin{equation}\label{equation_two}
S=Y_i + Y_{i+1} +Y_{i+2}+.....+Y_{i+count}
\end{equation}
Where, \emph{S} is the aggregate sum of sales over all products in that week for the category. The \emph{S} can be easily obtained from domain knowledge. \emph{$Y_i$} is the weekly sale of product 1, \emph{$Y_{i+1}$} is weekly sale of product 2 and so on. Here, \emph{count} represents the number of products in the given week in our dataset and can vary week over week.

For the week \emph{i}, \emph{$Y_i$} represents the sales of a high demand product that would suppress the weekly sales of other products accordingly. It is because, \emph{S} is approximately constant in a given week.
The idea is to exploit this useful information to make predictions for the cannibalized products. Even though the cannibalization information is not a part of input matrix \emph{X}, we construct a three XGBoost models given by \emph{$f_{1}(X)$}, \emph{$f_{2}(X)$}, \emph{$f_{3}(X)$}. Each of these models is trained on their corresponding objective functions \emph{$l_{1}$}, \emph{$l_{2}$}, \emph{$l_{3}$} respectively.
According to their usage in our framework they have been named as follows. XGBoost1 as baseline model, XGBoost2 as constraint model, XGBoost3 as fine-tune model as shown in flowchart \ref{FlowChart1}.

In stage 1, we train XGBoost on 1 to \emph{m} samples on SE objective function. We refer to this model as XGBoost1.
The SE objective function is given by,
\begin{equation}\label{equation_three}
\emph{$l_1$}=\frac{\sum_{i=1}^{m}(Y_{a_i} - Y_{p_i})^2}{m}.    
\end{equation}
Where $m$ is the number of training samples with actual sales. $Y_{a_i}$ is the weekly sale of an individual product. $Y_{p_i}$ is the predicted weekly sale by the model for an individual product. The trained XGBoost1 model is then used to predict for test set \emph{m}+1 to \emph{n} samples. The predictions on test set for samples \emph{m}+1 to \emph{n} in the dataset are used to update the columns \emph{sales(t-3)}, \emph{sales(t-2)},  \emph{sales(t-1)} and \emph{sales(t)}. Here, \emph{sales(t-3)}, \emph{sales(t-2)}, \emph{sales(t-1)} are a part of input feature set \emph{X}. This process of using previous week sales to make current prediction is called back-padding. The sales \emph{sales(t)} is the output label \emph{Y} that we are predicting.

The prediction made at stage 1 are of poor accuracy as it does not take into account the cannibalization of sales due to launch of new products. These predicted values $y_p$ are appended to $y_a$ column which initially only contained \emph{m} training samples. This updated train dataset containing \emph{n} samples is used as an input to stage 2 (stage 2 is explained in the next paragraph) of the algorithm. Also, XGBoost1 acts as a baseline model that gives the upcoming stages of the model a general guideline to make a better prediction.

In stage 2, we train XGBoost  on entire dataset containing \emph{n} samples based on the updated dataset obtained from stage 1. We refer to it as XGBoost2. Here, 1 to \emph{m} data-points contains actual sales and \emph{m}+1 to \emph{n} contain the predicted weekly sales by XGBoost1.
XGBoost2 is trained on the below objective function,
 \begin{equation}\label{equation_four}
\emph{$l_2$}=\frac{\sum_{i=1}^{n}(Y_{a_i} - Y_{p_i})^2}{n} + \frac{\sum_{i=1}^{n}(Y_{categoryActual_i}  -Y_{categoryPrediction_i} )^2}{count_i * n}
 \end{equation}
where $Y_{a_i}$ is the actual weekly sales of each product for i in [1,2,...m]. It also includes the predicted sales on test set by XGBoost1 model for i in [\emph{m+1}, \emph{m+2},...,\emph{n}]. $Y_{p_i}$ is the weekly sales for each product predicted by XGBoost2 model. $Y_{categoryActual_i}$ is the aggregate sum of sales of all products for a given week.
In equation \ref{equation_four} \emph{$Y_{categoryActual_i}$} is given by,
 \begin{equation}\label{equation_five}
 \emph{$Y_{categoryActual_i}$}=\sum_{i=1}^{count_i}Y_{a_i\; ; \; for \;i=1, 2, 3, ...., m.} 
  \end{equation}
Where, $Y_{categoryActual_i}$ is the aggregate sum of sales for a given category for week i and is assumed to be known. \emph{$Y_{categoryPrediction_i}$} in \ref{equation_four} is given by,
 \begin{equation}\label{equation_six}
\emph{$Y_{categoryPrediction_i}$}=\sum_{i=1}^{count_i}Y_{p_i} 
   \end{equation}
 where $count_i$ is the  number of products on sale in the given week \emph{i}. The XGBoost2 model 
makes predictions and updates the columns \emph{sales(t-3)},\emph{sales(t-2)}, \emph{sales(t-1)}  for records \emph{m+1} to \emph{n}. Here, we do not update the label column \emph{Y}, i.e. \emph{sale(t)}. We only update the input feature columns of \emph{X}.

As evident from equation \ref{equation_four}, the model in stage 2 is trained simultaneously on an objective function consisting of two parts namely, SE as well as categorical sum constraint. 
The $y_p$ from stage 1 that are appended to $y_a$ means that the new prediction tries to get close to $y_a$ as per SE. But it is forced to increase or decrease the prediction of all the products within the category for the week based on the categorical sum constraint part of the equation \ref{equation_four} which has been expanded in \ref{equation_five}. This ensures that if the products are over-forecasting or under forecasting then it self adjusts due to the sum constraint term in the objective function of equation \ref{equation_four}. 

In stage 3, we train XGBoost on entire dataset containing 1 to \emph{n} samples with the objective function given below and refer to it as XGBoost3.
 \begin{equation}\label{equation_seven}
 \emph{$l_3$}= \frac{\sum_{i=1}^{n}(Y_{categoryActual_i}  -Y_{categoryPrediction_i} )^2}{count_i * n} + \frac{\sum_{i=1}^{n}(Y_{p_i} - Y_{predRatio_i}*Y_{categoryActual_i})}{n}
 \end{equation}
where $Y_{predRatio_i}$  is obtained from output of stage 1. It is given by,
\begin{equation}\label{equation_eight}
Y_{predRatio_i}=\frac{Y_{p_i}}{Y_{categoryPrediction_i} }.
 \end{equation}

\subsection{Flow Chart}\label{FlowChart1}

\begin{tikzpicture}[node distance=2cm]
\node (start) [startstop, xshift=10cm] {Start};
\node (in1) [io, below of=start,text width=5 cm] {Train on historical data 1 to m records};
\node (pro1) [process, below of=in1] {XGBoost1};
\node (in2) [io, below of=pro1, text width =7 cm] {Append predictions m+1 to n and back pad to the dataset };
\node (in3) [io, right of=pro1, xshift=3.5cm,text width =5 cm, text height =0.3 cm] {Generated product prediction ratio };
\node (pro2) [process, below of=in2] {XGBoost2};
\node (in4) [io, below of=pro2,text width = 7 cm] {Generate predictions and only back pad them to dataset };
\node (pro3) [process, below of=in4] {XGBoost3};
\node (in5) [io, below of=pro3,] {Final prediction};
\node (Stop) [startstop, below of=in5] {Stop};

\draw [arrow] (start) -- (in1);
\draw [arrow] (in1) -- (pro1);
\draw [arrow] (pro1) -- (in2);
\draw [arrow] (pro1) -- (in3);
\draw [arrow] (in3) |- (in5);
\draw [arrow] (in2) -- (pro2);
\draw [arrow] (pro2) -- (in4);
\draw [arrow] (in4) -- (pro3);
\draw [arrow] (pro3) -- (in5);
\draw [arrow] (in5) -- (Stop);
\end{tikzpicture}
\\*\-\hspace{1cm} The algorithm consists of three stages, with each stage utilizing the information from previous stages and gets trained on a different objective function.
In stage 2 the model increased or decreased the predictions of all products based on categorical sum. Stage 3, is to guide the prediction of each product within the category based on equation \ref{equation_seven}. To do this we use the feature device\_prediction\_ratio, which provides information of how much sale is contributed by individual device within category sum $S$.
The objective function in stage 3 is given by equation \ref{equation_seven} and consists of two terms. The first term is same as sum constraint in equation \ref{equation_four}. This ensures that the sum of  predictions of all the devices in the category is still adhering to the category actuals obtained from domain knowledge. The second term in the objective function is to make sure that individual device sale prediction within category increases or decreases based on the information obtained from stage 1, respectively.
The stage 3 is the final stage in our framework that insures that the final prediction obtained for each device sums upto the categorical sum obtained from domain knowledge and also adjusting the individual device level predictions according to their input features.

\section{Experimentation and Results}

Forecasting experiment has been carried out on A, B and C categories. Each category has set of devices where forecasting has been done upto 14 weeks.
The experiments have been performed on devices of A, B and C categories once NPI devices have been launched in their respective category.
\\*\-\hspace{1cm} Table \ref{tab:Table_two} enlists the details about the devices and the experiments performed. 
The \emph{category} column lists the devices belonging to a certain category based on their pricing and device features. As discussed above in equation \ref{equation_two}, it is considered that total sales volume of each category is known. The objective is to predict the sales for existing devices of each category.
The \emph{Date} column tells us the starting date of the forecasts made by the models.
\emph{Regular XGBoost Accuracy} column is the weighted accuracy over the entire forecasting horizon of 8 weeks or 14 weeks depending on the category based on the prediction made by XGBoost using SE as the objective function in equation \ref{equation_three}.
Similarly, the \emph{Three-Stage XGBoost} column is the weighted accuracy over the entire forecasting horizon for the given category based on the prediction made by our three-stage XGBoost framework given by equation \ref{equation_seven}.
\emph{Lags} represent the number of weeks from the staring date for which forecasting has been carried out.
Existing Devices are the devices for which sales forecasting has been carried out. Sales for these devices are impacted by launch of new devices belonging to the same category. 
New devices are the devices that cause cannibalization of existing devices. The forecasting of new devices is out of scope for this study and is a good direction for future research.
\\*\-\hspace{1cm} The train data consists of historical data of about 4 years depending on the experimental setup. We have used weighted accuracy to calculate the performance of the models over different experiments. The weighted accuracy is given by,
 \begin{equation}\label{equation_nine}
\\*Weighted Accuracy= 1- \frac{\sum_{i=1}^{m}abs(y\_actual-y\_predict)}{sum}.
 \end{equation}
 \\*Where, $sum=\sum_{i=1}^{m}y\_actual$.
The important input features to the model are indicators to showcase holiday seasonality, promotional features, device specifications and other features such as weeks from device launch and pricing, etc.

\begin{figure}[H]
\centering
\parbox{6cm}{
\includegraphics[width=5.5cm]{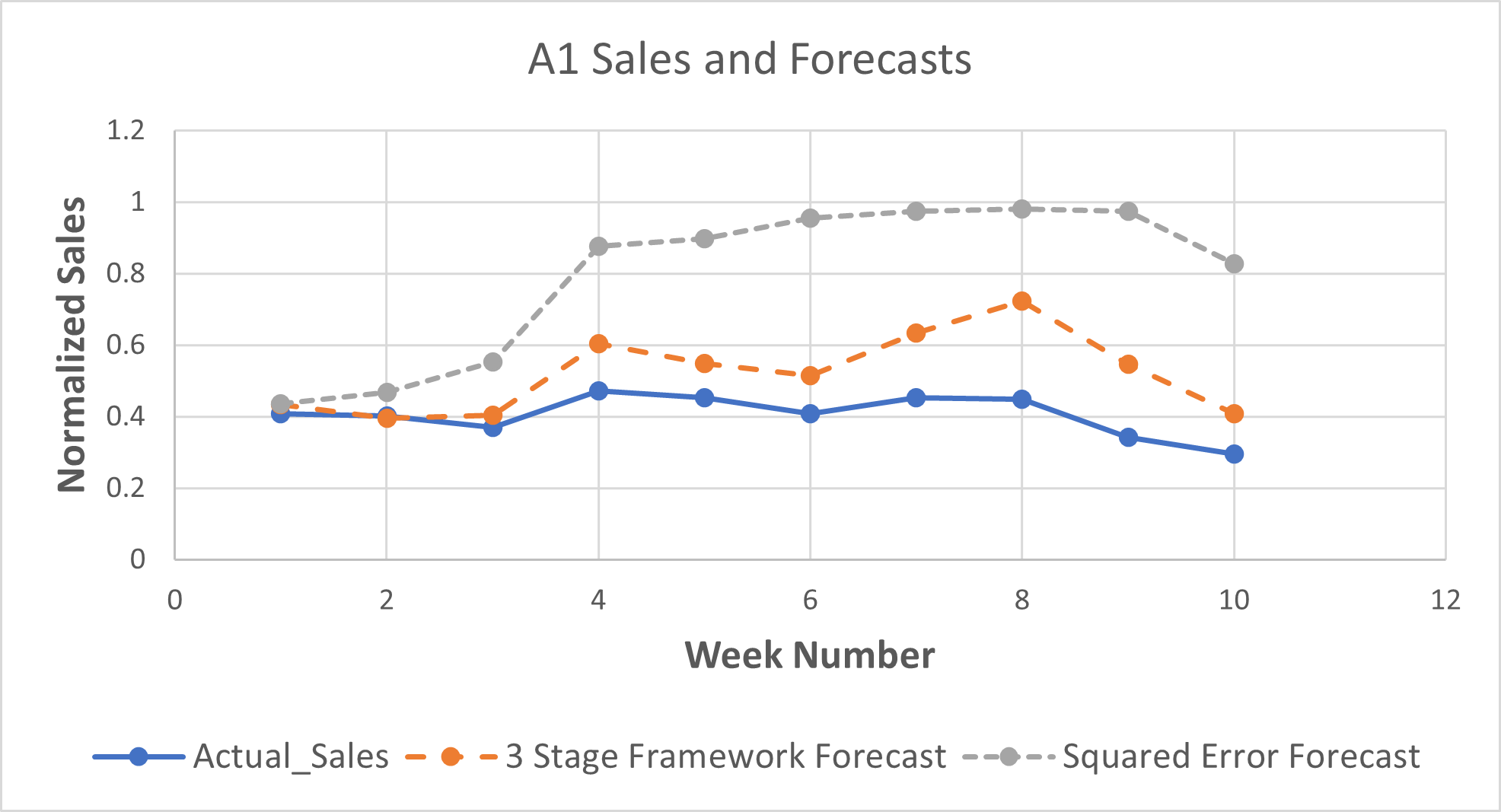}
\caption{A1 Sales and Forecasts}
\label{fig:A1}}
\qquad
\end{figure}

For the figures in this section, the horizontal axis represents the time horizon over which sales and forecast have been plotted. The vertical axis represents the corresponding actual sales and forecasts generated by the model, the sales and forecasts have been normalized. Figure \ref{fig:A1} represents the weekly normalized sales belonging to category A. We see that three-stage forecast from XGBoost model is much more closer to actual sales compared to SE based XGBoost forecast. We also observe that after week 8 there is reduction in actual sales as well as in the forecast produced by three-stage XGBoost model. However, the SE XGBoost model is unable to capture drop in sales correctly and thus we see that long term forecast has been accurately captured.

We can observe that prediction with our framework yields better results compared to existing machine learning model. It is because it bears a much closer resemblance with the actual trend.
We have identified week 3 and week 4 as Thanksgiving dates where we expect to see a spike in sales.
Similarly, the newer model has the ability to capture spike on week 8 which is the Christmas week as seen in figure \ref{fig:A1}.

\begin{figure}[H]
\centering
\parbox{5cm}{
\includegraphics[width=5.5cm]{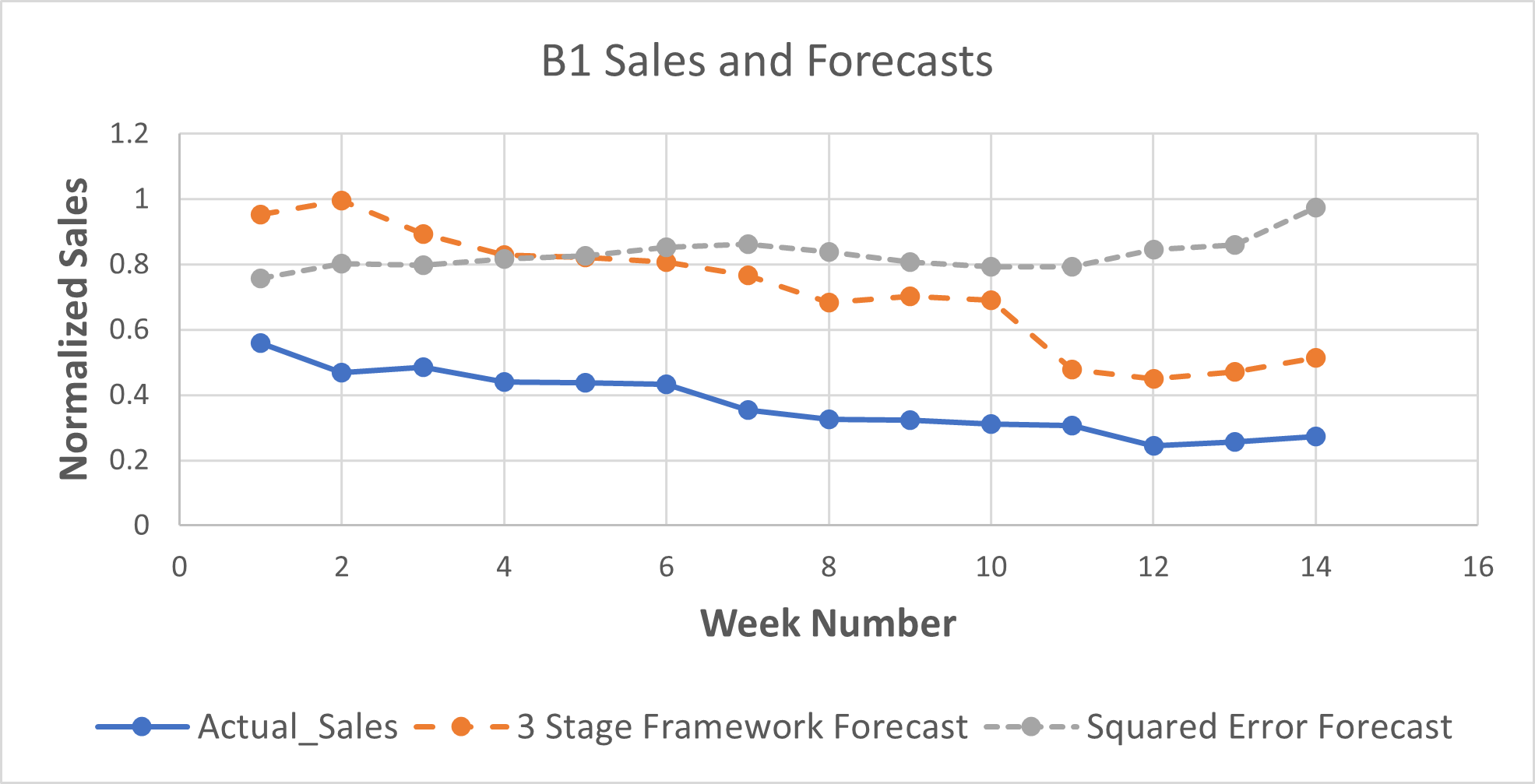}
\caption{B1 Sales and Forecasts}
\label{fig:B1}}
\qquad
\begin{minipage}{5cm}
\includegraphics[width=5.5cm]{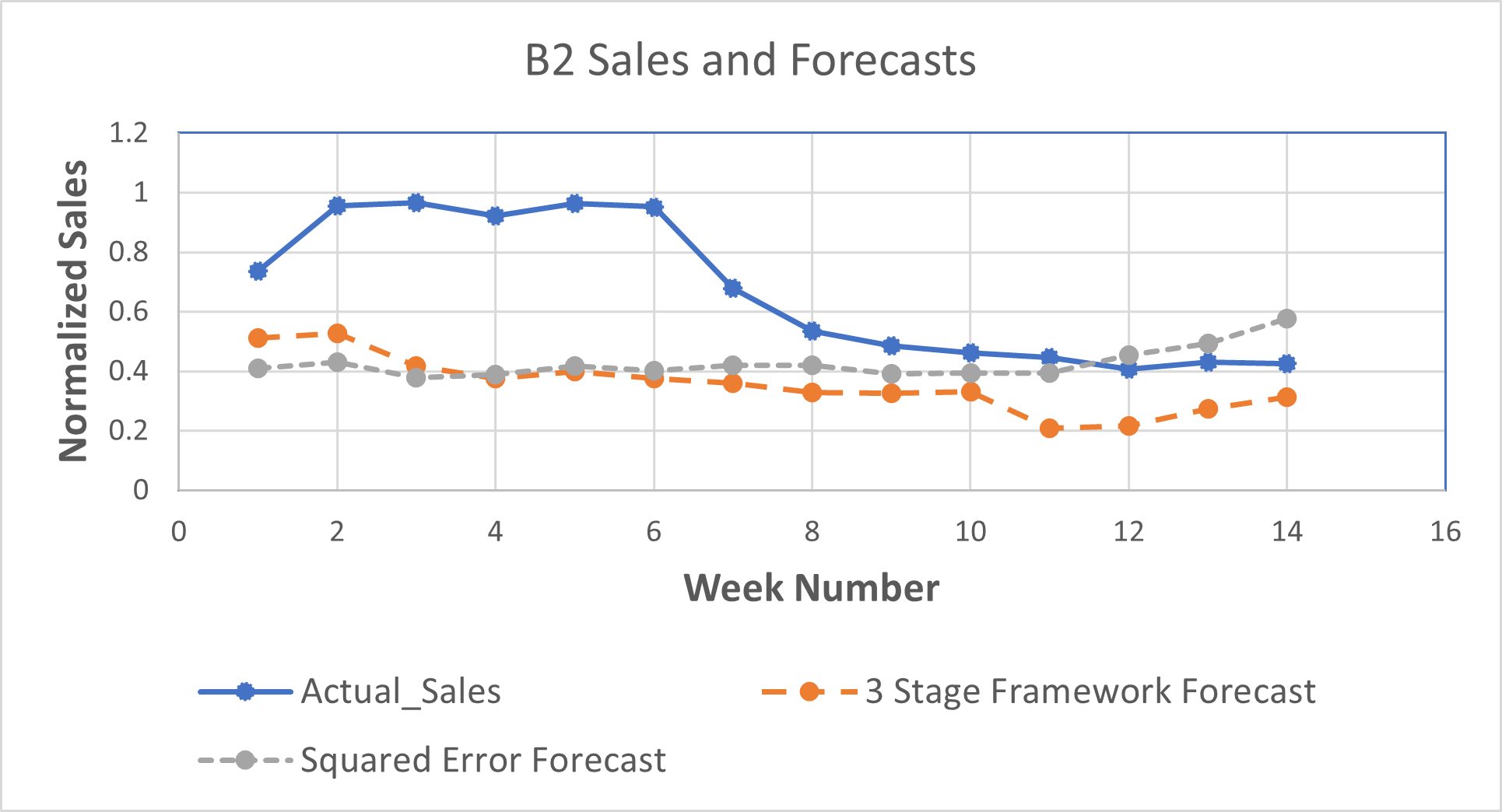}
\caption{B2 Sales and Forecasts}
\label{fig:B2}
\end{minipage}
\end{figure}

Figures \ref{fig:B1} and \ref{fig:B2} represent the results for category\_B. Actual sales in B1 gradually decreases over 14 week horizon.
The reason being, NPI devices B11, B12, B13 are launched on Aug 17, 2020 respectively. However, SE objective function based XGBoost model is unable to capture the same as it produces an increase in forecast instead of decreasing. The three-stage XGBoost model captures the dynamics very well especially over longer time horizon. More importantly, the performance of the three-stage framework outperforms the performance of SE XGBoost after 5th week.

\begin{figure}[H]
\centering
\parbox{5cm}{
\includegraphics[width=5.5cm]{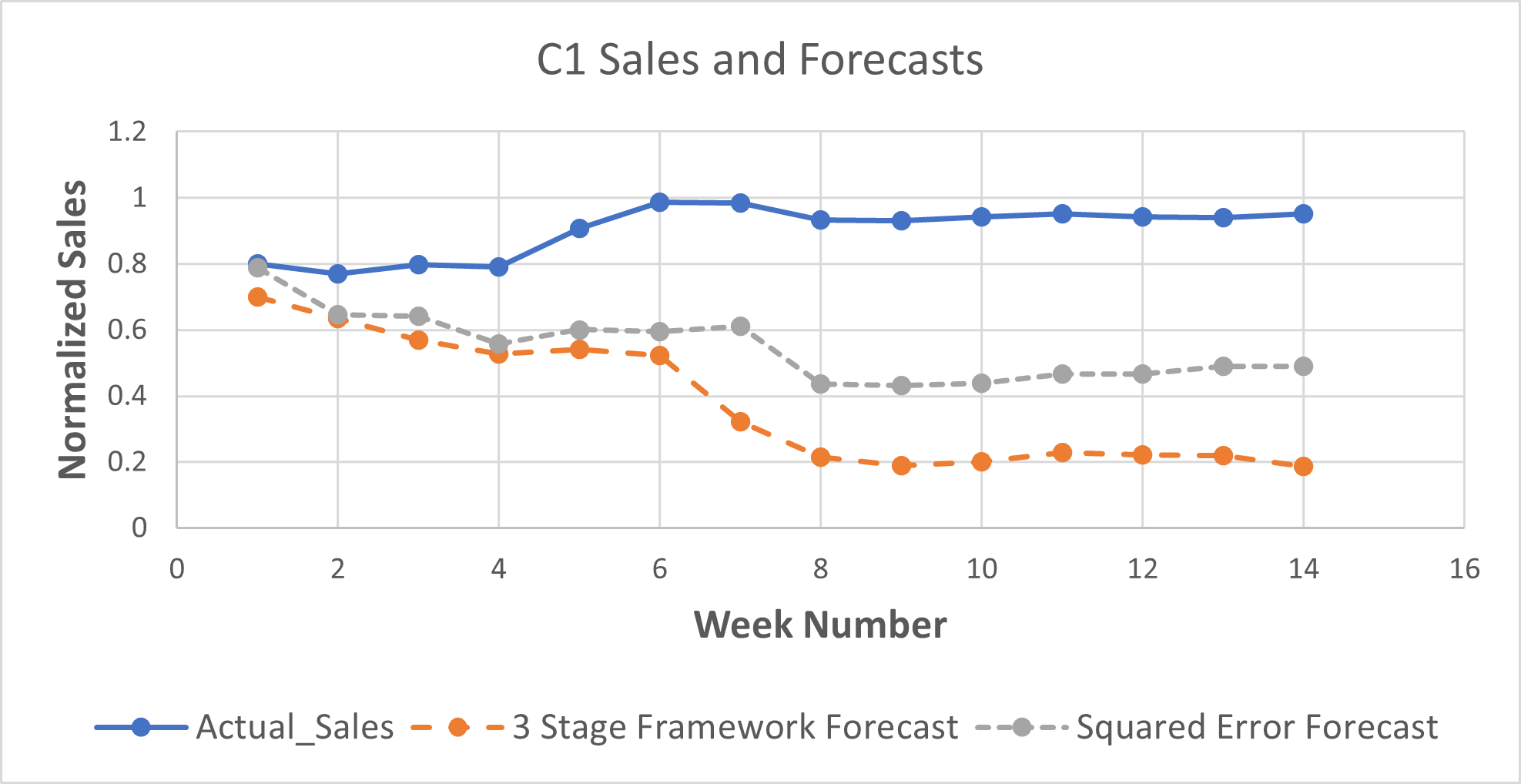}
\caption{C1 Sales and Forecasts}
\label{fig:C1}}
\qquad
\begin{minipage}{5cm}
\includegraphics[width=5.5cm]{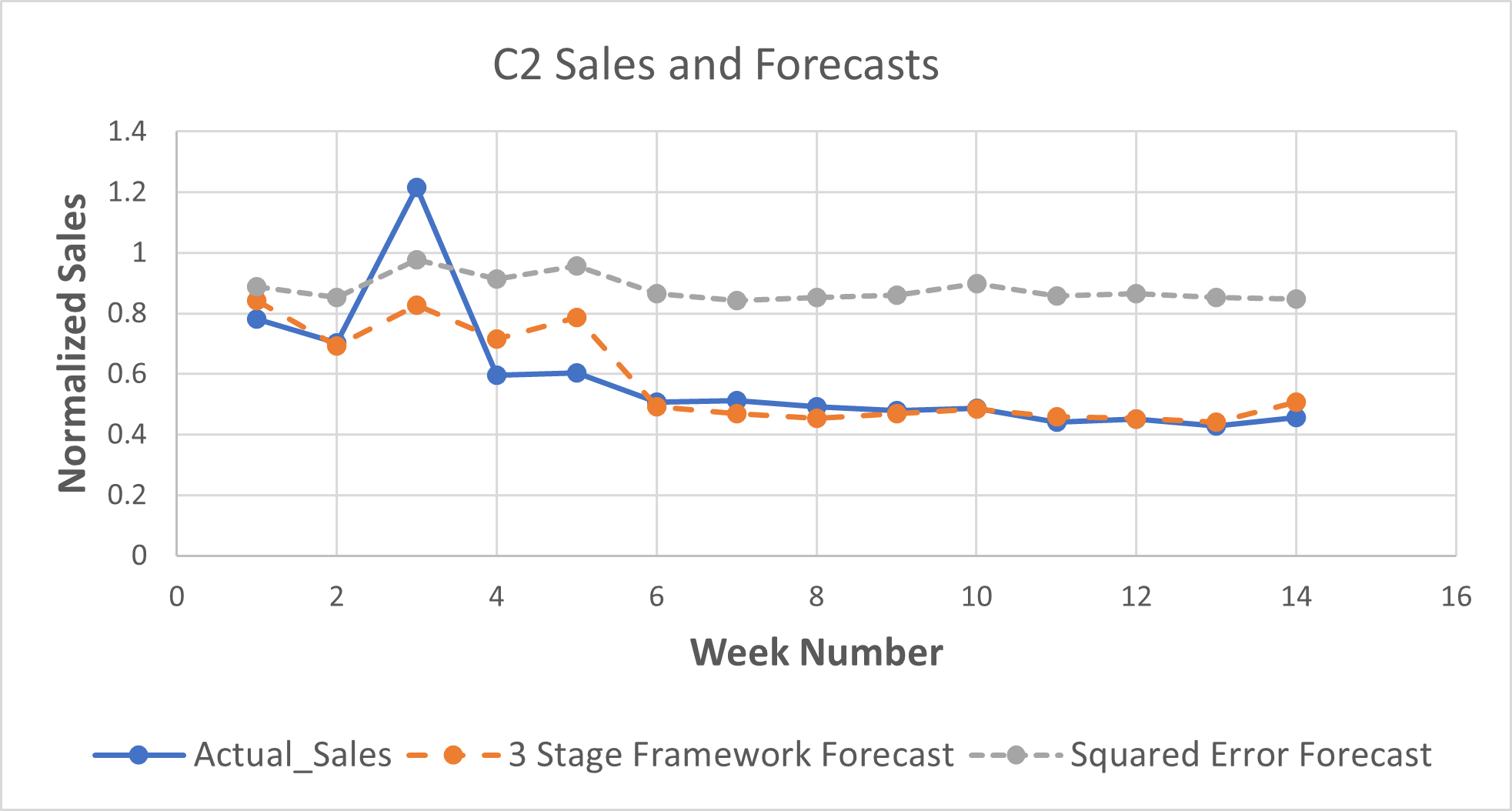}
\caption{C2 Sales and Forecasts}
\label{fig:C2}
\end{minipage}
\end{figure}

Actual and forecasts for devices C1 and C2 belonging to  category\_C are shown in figure \ref{fig:C1} and \ref{fig:C2}, respectively. We see that overall accuracy is greater in the three-stage algorithm compared to standard SE objective function based XGBoost.
We observe that for product C1 the SE and the three-stage XGBoost framework both perform poorly in capturing the market dynamics, especially after week 7. Both models tend to significantly under-forecast the sales after week 7 compared to actuals.
For product C2 belonging to category C we see three-stage XGBoost model performs much better compared to SE XGBoost model. The three-stage XGBoost for C3 product has forecasts very close to actuals. On the contrary, the SE XGBoost performs quite poorly.
It can be observed, the three-stage framework based XGBoost consistently performs better than SE based XGBoost in all of our experiments.
\\*\-\hspace{1cm} It can be seen that our model exploits categorical sales that is a input to the three-stage framework. Hence, the three-stage framework uses the total categorical sale to adjust the sales of individual devices within the
category. This helps to overcome the error propagation problem when making long term forecasts.
Hence, in all the above cases we observe that the proposed three-stage framework consistently outperforms existing state of the art XGBoost model significantly.
The performance of the models has been calculated by using the weighted accuracy given by equation \ref{equation_nine}.

\begin{table}[htp!]
\caption {Experiment Results} \label{tab:Table_two} 
\centering
\begin{center}
\scalebox{0.8}{
\begin{tabular}{|l|l|l|l|l|l|l|}
\hline
 \begin{tabular}[c]{@{}l@{}} Category\end{tabular}
 &\begin{tabular}[c]{@{}l@{}} Date\end{tabular}
  &\begin{tabular}[c]{@{}l@{}} Regular XGBoost\\ Accuracy\end{tabular}
   &\begin{tabular}[c]{@{}l@{}} Three-Stage\\ XGBoost\end{tabular}
   &\begin{tabular}[c]{@{}l@{}} Lags\end{tabular}
   &\begin{tabular}[c]{@{}l@{}} Existing Devices\end{tabular}
   &\begin{tabular}[c]{@{}l@{}} New Devices\end{tabular}
  \\ \hline Category\_A & 02 Nov, 2020 & 38.65\%  & 67.09\% & 8&\begin{tabular}[c]{@{}l@{}}A1, A2, A3,\\ A4, A5\end{tabular} & \begin{tabular}[c]{@{}l@{}}A6, A7, A8, A9\end{tabular}
    \\ \hline Category\_B & 17 Aug, 2020 & 44.60\%  & 51.50\% & 14&\begin{tabular}[c]{@{}l@{}}B1, B2, B3, B4, B5, B6,\\ B7, B8, B9, B10 \end{tabular} & \begin{tabular}[c]{@{}l@{}} B11, B12, B13\end{tabular}
     \\ \hline Category\_C & 17 Aug, 2020 & 6.70\%  & 52.90\% & 14&\begin{tabular}[c]{@{}l@{}}C1, C2, C3, C4,\\ C5, C6, C7, C8, C9, C10, C11 \end{tabular} & \begin{tabular}[c]{@{}l@{}} C12, C13, C14 \end{tabular}\\ \hline

\end{tabular}}
\end{center}
\end{table}

\section{Conclusion}
In this work, we have developed an algorithm to improve the sales forecasting accuracy of older devices that are impacted by cannibalization due to launch of new devices. The other problem statement that has been addressed is to improve week over week long term forecasting accuracy of the old devices.
To address the above two issues we developed a three-stage framework using XGBoost algorithm that consists of three stages.
We compared the three-stage XGBoost-based framework with the regular XGBoost model that uses SE as the objective function.
Our experiments show that the proposed three-stage framework performs consistently better for long term forecasts.
We used weighted accuracy as a metric to quantify the performance comparison.
From our experiments, we observe a significant increase in overall prediction accuracy for old products from 38\% in the baseline model, to 67\% after using the proposed framework.
A good direction for future work is to extend the framework to make accurate forecasts for newly launched devices that cause the cannibalization of sales of old devices.

\nocite{*}
\bibliographystyle{plain}
\bibliography{interactapasample.bib}
\end{document}